\PassOptionsToPackage{numbers,sort&compress}{natbib}
\documentclass{article}
\usepackage[preprint]{neurips_2026}
\usepackage[T1]{fontenc}
\usepackage[utf8]{inputenc}
\usepackage{lmodern}
\usepackage{microtype}
\usepackage{graphicx}
\usepackage{booktabs}
\usepackage{array}
\usepackage{amsmath,amssymb}
\usepackage{hyperref}
\usepackage[inline]{enumitem}
\usepackage[dvipsnames]{xcolor}
\usepackage{caption}
\usepackage{tikz}
\usetikzlibrary{shapes.geometric,arrows.meta,positioning,fit,backgrounds}

\newcommand{\coreSamples}{16{,}783}
\newcommand{\expansionSamples}{48{,}072}
\newcommand{\matchedExpansionSamples}{44{,}066}
\newcommand{\productionProbeSamples}{4{,}006}

\newcommand{\neurips}{48{,}072}

\newcommand{\corePairs}{3}

\newcommand{\coreByteIdentity}{90.66\%}
\newcommand{\coreByteIdentityN}{2{,}592 / 2{,}859}

\newcommand{\coreTOSTpass}{25/27}

\newcommand{\seventyBrefusal}{0.839}
\newcommand{\seventyBrefusalCI}{[0.809, 0.864]}
\newcommand{\seventyBPhaseTwoThreeRefusal}{0.840}
\newcommand{\seventyBPhaseTwoThreeRefusalCI}{[0.783, 0.884]}
\newcommand{\seventyBAdvBenchN}{700}
\newcommand{\seventyBPhaseTwoThreeN}{200}
\newcommand{\maxCohensH}{0.024}
\newcommand{\nullCutoffH}{0.1}
\newcommand{\bfsixteenShiftLow}{36\%}
\newcommand{\bfsixteenShiftHigh}{53\%}
\newcommand{\expansionIdentity}{100.00\%}
\newcommand{\efourSeedCount}{12{,}018}

\newcommand{\holmAlpha}{0.0045}
\newcommand{\tostBand}{$\pm 3$pp}

\newcommand{\heldoutNcal}{10}
\newcommand{\heldoutNeval}{7}
\newcommand{\heldoutMaxHCal}{0.0243}
\newcommand{\heldoutHstarCal}{0.097}
\newcommand{\heldoutMaxHEval}{0.000}

\newcommand{\tais}{TAIS}
\newcommand{\taisLong}{Typical-Acceptance Invariance Screen}

\graphicspath{{figures/}}
\hypersetup{
  pdfkeywords={speculative decoding, LLM serving, safety evaluation, behavioral equivalence, TOST},
  colorlinks=true,
  linkcolor=MidnightBlue,
  citecolor=MidnightBlue,
  urlcolor=MidnightBlue,
  pdfauthor={Sahil Kadadekar},
  pdftitle={Speculative Decoding at Temperature Zero: A Scoped Safety-Invariance Screen with a 48,072-Sample Expansion},
  pdfsubject={Preprint},
  pdfcreator={LaTeX with hyperref},
  pdfproducer={}
}

\title{Speculative Decoding at Temperature Zero:\\A Scoped Safety-Invariance Screen with a \neurips{}-Sample Expansion}
\author{Sahil Kadadekar\\
Independent Researcher\\
\texttt{sahilkadadekar@nyu.edu}}
\date{}

\begin{document}

\maketitle

\begin{abstract}
Speculative decoding accelerates inference by letting a draft model propose tokens for a target model to verify, raising a concrete safety question: at temperature zero, can draft-side behavior leak into safety-scored outputs? We answer with \taisLong{} (\tais{}), a behavioral-equivalence screen that pairs target-only and speculative outputs on the same safety battery and requires byte-identity evidence, TOST equivalence at \tostBand{}, and per-task Cohen's~$h$ below a calibrated null cutoff of $|h|<\nullCutoffH{}$. Applied to a \coreSamples{}-sample confirmatory core plus \matchedExpansionSamples{} matched expansion samples (fp16/bf16 execution, canonical and DPO-adversarial drafts, GPTQ-4bit drafts, two seeds, and four safety benchmarks), the tested temperature-zero vLLM stacks show no detectable safety divergence under \tais{}. The largest absolute Cohen's~$h$ on matched target-only versus speculative refusal is \maxCohensH{}, roughly an order of magnitude below the conventional trivial-effect floor; 25 of 27 per-task TOST contrasts pass at the \tostBand{} margin (the two non-pass contrasts are capability-domain Wald-CI edge cases at $p_{\text{target}}{=}p_{\text{spec}}$ at the ceiling, not genuine non-equivalence); the DPO-adversarial draft produces byte-identical output to the canonical draft across 4{,}006 samples (the expected algorithmic consequence of strict rejection sampling at $T{=}0$, validated operationally rather than presented as discovery); and bf16 changes \bfsixteenShiftLow{}--\bfsixteenShiftHigh{} of output bytes without moving any per-task safety rate outside equivalence. A separate \productionProbeSamples{}-sample 70B production-scale probe, which lacks a matched 70B target-only arm and is therefore not counted as a \tais{} pass, produces AdvBench refusal \seventyBrefusal{} over \seventyBAdvBenchN{} AdvBench completions with 95\% Wilson CI \seventyBrefusalCI{}. We make no claim about sampling temperatures, untested frameworks, untested model families, or tree-speculation variants such as EAGLE and Medusa.
\end{abstract}

\section{Introduction}
\label{sec:intro}

Safety can move at several points in the model pipeline. Training-time alignment can trade refusal behavior against other capabilities \citep{Huang2025SafetyTax}, and even benign or adversarial fine-tuning can quickly erode aligned behavior \citep{Qi2023Finetune}. Deployment-time compression is another axis: quantization and related methods can change refusal, truthfulness, and bias behavior even when quality metrics remain stable. This paper studies a third axis: inference-time acceleration.

Speculative decoding~\citep{Leviathan2023Speculative,Chen2023Accelerating,Stern2018Blockwise} is now common in vLLM~\citep{Kwon2023PagedAttention}, TensorRT-LLM, SGLang, and HuggingFace TGI, and follow-on work has produced tree variants~\citep{Cai2024Medusa,Li2024EAGLE,Miao2024SpecInfer,Chen2024Sequoia}, retrieval-based drafting~\citep{He2024REST}, and parallel/lookahead schemes~\citep{Fu2024Lookahead}; \citet{Xia2024SpecSurvey} survey the design space. A smaller draft model proposes tokens and a larger target verifies them. Because the draft has its own alignment profile, it is natural to ask whether a weaker or adversarial draft can push unsafe behavior through the acceptance pathway. The closest published security-side result on speculative decoding is \citet{Wei2024SideChannels}, who exhibit a timing side-channel that lets an observer fingerprint inputs and exfiltrate datastore contents from speculative-decoding traffic; that work demonstrates speculative decoding as a security-relevant surface, but it does not test whether draft-side alignment changes verified safety-scored outputs. Whether the acceptance pathway itself can leak draft-side behaviour into safety-scored verified outputs is the question this paper takes up empirically.

We test the greedy operating point. At temperature zero, does speculative decoding change safety-scored outputs relative to target-only decoding on the same inputs? Our answer is a bounded null: in the tested matched vLLM configurations, across Llama and Qwen model families, four safety benchmarks, canonical and adversarial drafts, GPTQ-4bit drafts, two seeds, and fp16/bf16 execution, greedy speculative decoding shows no detectable safety divergence from target-only decoding under the screen defined below. A separate 70B run tests whether the same production-scale speculative stack remains in the expected refusal range, but because it does not include a matched 70B target-only arm, it is reported as a scale sanity probe rather than as a \tais{} equivalence cell.

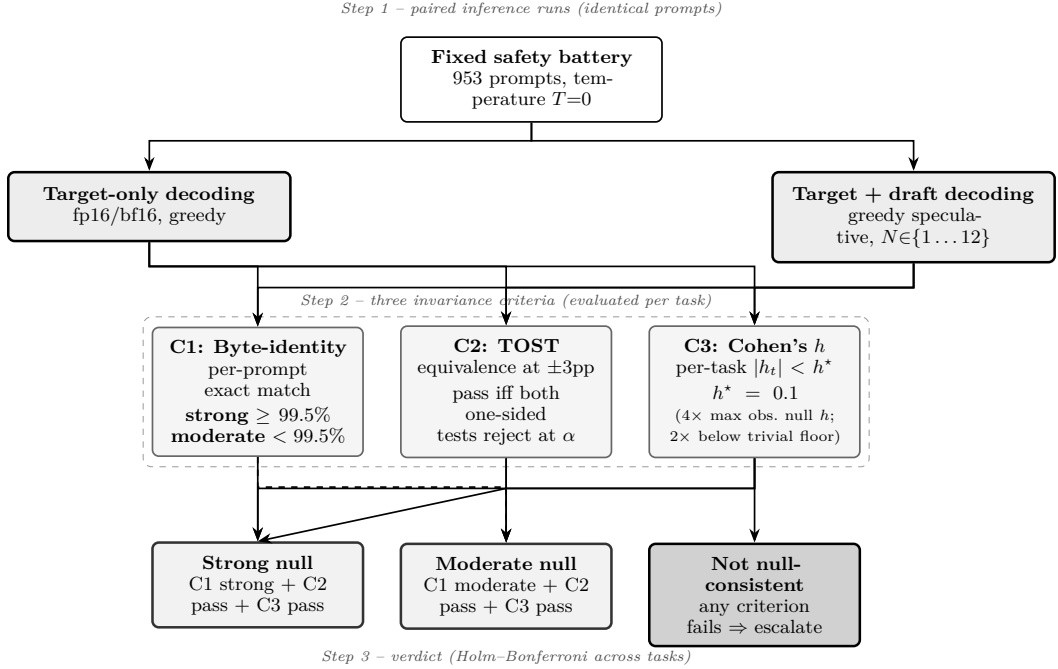
\begin{figure}[t]
\centering
\resizebox{\linewidth}{!}{%
\begin{tikzpicture}[
  >=Stealth,
  font=\small,
  node distance=4mm and 10mm,
  %--- node styles ---
  inputbox/.style={
    rectangle, rounded corners=3pt,
    draw=black, thick,
    text width=0.28\textwidth, align=center,
    inner sep=6pt, minimum height=2em
  },
  stepbox/.style={
    rectangle, rounded corners=3pt,
    draw=black, very thick,
    fill=black!7,
    text width=0.30\textwidth, align=center,
    inner sep=7pt, minimum height=2.4em
  },
  critbox/.style={
    rectangle, rounded corners=3pt,
    draw=black!60, thick,
    fill=black!4,
    text width=0.22\textwidth, align=center,
    inner sep=6pt, minimum height=4.2em
  },
  verdictbox/.style={
    rectangle, rounded corners=3pt,
    draw=black, very thick,
    fill=black!12,
    text width=0.22\textwidth, align=center,
    inner sep=6pt, minimum height=2.2em
  },
  labelstyle/.style={font=\scriptsize\itshape, text=black!70}
]

%--- Row 0: shared input ---
\node[inputbox] (prompts)
  {\textbf{Fixed safety battery}\\953 prompts, temperature~$T{=}0$};

%--- Row 1: two parallel decode streams ---
\node[stepbox, below left=8mm and 18mm of prompts] (target)
  {\textbf{Target-only decoding}\\fp16/bf16, greedy};
\node[stepbox, below right=8mm and 18mm of prompts] (spec)
  {\textbf{Target + draft decoding}\\greedy speculative, $N{\in}\{1\dots12\}$};

% arrows from prompts
\draw[->, thick] (prompts.south) -- ++(0,-3mm) -| (target.north);
\draw[->, thick] (prompts.south) -- ++(0,-3mm) -| (spec.north);

%--- Row 2: three criteria boxes ---
\node[critbox, below=14mm of target, xshift=18mm] (byte)
  {\textbf{C1: Byte-identity}\\
   per-prompt exact match\\[2pt]
   \textbf{strong} $\geq 99.5\%$\\
   \textbf{moderate} $< 99.5\%$};

\node[critbox, right=6mm of byte] (tost)
  {\textbf{C2: TOST}\\
   equivalence at $\pm 3$pp\\[2pt]
   pass iff both one-sided\\
   tests reject at $\alpha$};

\node[critbox, right=6mm of tost] (cohenh)
  {\textbf{C3: Cohen's $h$}\\
   per-task $|h_t| < h^\star$\\[2pt]
   $h^\star = 0.1$\\
   {\scriptsize(4$\times$ max obs.\ null $h$;\ 2$\times$ below trivial floor)}};

% horizontal bracket / grouping background
\begin{scope}[on background layer]
  \node[draw=black!40, dashed, rounded corners=4pt,
        inner sep=4pt, fit=(byte)(tost)(cohenh),
        label={[labelstyle]above:\textit{Step 2 -- three invariance criteria (evaluated per task)}}] {};
\end{scope}

% arrows from decode outputs into criteria boxes
\draw[->, thick] (target.south) -- ++(0,-4mm) -| (byte.north);
\draw[->, thick] (target.south) -- ++(0,-4mm) -| (tost.north);
\draw[->, thick] (target.south) -- ++(0,-4mm) -| (cohenh.north);
\draw[->, thick] (spec.south)   -- ++(0,-4mm) -| (byte.north);
\draw[->, thick] (spec.south)   -- ++(0,-4mm) -| (tost.north);
\draw[->, thick] (spec.south)   -- ++(0,-4mm) -| (cohenh.north);

%--- Row 3: verdicts ---
\node[verdictbox, below=14mm of byte,
      fill=black!5, draw=black!80] (strong)
  {\textbf{Strong null}\\C1 strong + C2 pass + C3 pass};

\node[verdictbox, below=14mm of tost,
      fill=black!5, draw=black!80] (moderate)
  {\textbf{Moderate null}\\C1 moderate + C2 pass + C3 pass};

\node[verdictbox, below=14mm of cohenh,
      fill=black!18, draw=black] (nonnull)
  {\textbf{Not null-consistent}\\any criterion fails $\Rightarrow$ escalate};

% arrows from criteria to verdicts
\draw[->, thick] (byte.south)    -- ++(0,-5mm) -| (strong.north);
\draw[->, thick] (tost.south)    -- ++(0,-5mm) -- (strong.north);
\draw[->, thick] (cohenh.south)  -- ++(0,-5mm) -| (strong.north);

\draw[->, thick, dashed] (byte.south)   -- ++(0,-5mm) -| (moderate.north);
\draw[->, thick] (tost.south)   -- ++(0,-5mm) -- (moderate.north);
\draw[->, thick] (cohenh.south) -- ++(0,-5mm) -| (moderate.north);

\draw[->, thick] (cohenh.south) -- ++(0,-5mm) -| (nonnull.north);

% Step labels
\node[labelstyle, above=2mm of prompts] {\textit{Step 1 -- paired inference runs (identical prompts)}};
\node[labelstyle, below=2mm of moderate.south]
  {\textit{Step 3 -- verdict (Holm--Bonferroni across tasks)}};

\end{tikzpicture}
}
\caption{The \tais{} screen pairs target-only and speculative outputs on a fixed safety battery and requires three criteria to declare a null: byte-identity $\geq 99.5\%$ (strong) or lower (moderate), TOST equivalence at $\pm 3$pp, and per-task $|h|$ below the calibrated cutoff $h^\star = 0.1$.
The cutoff $h^\star = 0.1$ sits a factor of four above the loudest observed null-consistent contrast in this study ($|h| = 0.024$) and a factor of two below the Cohen ``trivial effect'' floor of 0.2; future applications should re-calibrate on new evidence.
Multiple comparisons across tasks are controlled with Holm--Bonferroni; a stack is null-consistent over the full battery only if every well-conditioned per-task contrast passes criteria C2 and C3. \emph{Wald-CI edge cases excluded:} per-task contrasts where $p_{\text{target}}{=}p_{\text{spec}}$ at a boundary (0.0 or 1.0), so the observed difference is exactly zero but the Wald CI degenerates, are flagged separately rather than counted as escalations to ``Not null-consistent.''}
\label{fig:tais-screen}
\end{figure}

\paragraph{Named method: \tais{}.}
We define the \emph{\taisLong{}} (\tais{}) as the behavioral-equivalence screen used to establish the null (see Figure~\ref{fig:tais-screen} for a protocol overview). Given a speculative-decoding stack and a matched safety battery, \tais{} pairs target-only and speculative outputs on identical inputs, measures byte-identity, evaluates TOST equivalence at \tostBand{}, and computes per-task Cohen's~$h$. A stack is null-consistent on the tested battery if $\max_t |h_t|$ is below \nullCutoffH{} and every well-conditioned per-task TOST contrast passes (capability-domain Wald-CI edge cases at the ceiling, where $p_{\text{target}}{=}p_{\text{spec}}$, are flagged separately rather than counted as failures). The cutoff is calibrated from this expansion: it sits above the loudest observed null-consistent contrast and below the conventional trivial-effect floor \citep{Cohen1988Stats}. \tais{} composes paired testing, TOST~\citep{Lakens2017TOST}, and Cohen's~$h$ into a calibrated equivalence protocol with explicit pass criteria.

\paragraph{Evidence scope.}
The confirmatory core contains \coreSamples{} paired samples on three Llama and Qwen pairs. The expansion adds \neurips{} samples across five probes: E1 scales to a Llama-3.1-70B target with an 8B draft as a one-arm production-scale probe; E2 inserts a DPO-adversarial draft trained on flipped Anthropic/hh-rlhf preferences; E3 swaps in a GPTQ-4bit draft; E4 repeats the core pairs over two seeds; and E5 switches the accumulator dtype from fp16 to bf16. Across the matched \tais{} probes, every well-conditioned per-task contrast falls inside the \tostBand{} equivalence bound (25 of 27 in the core ledger; the two non-passing contrasts are capability-domain Wald-CI edge cases at the ceiling), and the largest Cohen's~$h$ observed across matched AdvBench refusal contrasts is \maxCohensH{}.

\paragraph{Contributions.}
\begin{enumerate}[leftmargin=*]
  \item A scoped behavioral null: in the tested temperature-zero setting, greedy speculative decoding shows no detectable safety divergence from target-only decoding under byte-identity, TOST, and Cohen's~$h$ criteria across a \coreSamples{}-sample core and \matchedExpansionSamples{} matched expansion samples, with an additional \productionProbeSamples{}-sample 70B scale probe reported separately.
  \item \tais{}, a reusable equivalence screen with explicit pass criteria and a stated calibration procedure.
  \item A direct stress test of the acceptance-pathway leakage concern at the greedy operating point: a DPO-adversarial draft and a GPTQ-4bit draft produce no leakage detectable above the per-cell MDE under the target's verification step. We frame the byte-identity outcome on the DPO-adversarial draft as \emph{operational validation} that the \citet{Leviathan2023Speculative} distributional guarantee, which collapses to byte-equality at $T{=}0$ modulo numerical non-associativity, survives in practice on a strongly-perturbed draft, rather than as a novel theoretical discovery.
  \item A scope boundary that separates inference-time acceleration from training-time alignment and deployment-time compression: the null applies to temperature zero, vLLM v0.19, two model families, strict rejection sampling plus typical acceptance, and the tested safety battery (full enumeration in \S\ref{sec:discussion}).
\end{enumerate}

\section{Related Work}
\label{sec:related}

\paragraph{Safety tax across the pipeline.}
Modern safety alignment is built on RLHF and preference-learning
machinery~\citep{Christiano2017DeepRLHF,Ouyang2022InstructGPT,Bai2022Anthropic,Rafailov2023DPO},
yet the resulting alignment is fragile in identifiable ways.
\citet{Huang2025SafetyTax} identify a training-time \emph{safety tax}:
adding safety alignment to a reasoning-trained large model reduces
reasoning benchmark performance. \citet{Qi2023Finetune} make the
analogous finding from the fine-tuning side: even benign task-specific
fine-tuning on a small set of examples can erode the safety behaviour of
an aligned base model, and a separate jailbreak-attack literature shows
the same fragility from the input
side~\citep{Wei2023Jailbroken,Zou2023AdvBench,Kang2023Programmatic}.
Both establish that safety alignment is not a free property of the
weights. The present paper asks the same question about the
inference-time lever practitioners actually pull at deployment time:
speculative decoding.

\paragraph{Speculative decoding: theory and systems.}
Speculative decoding was introduced independently by
\citet{Leviathan2023Speculative} and \citet{Chen2023Accelerating},
building on earlier blockwise parallel
decoding~\citep{Stern2018Blockwise}; rejection sampling preserves the
target distribution exactly, while typical acceptance relaxes the
verification criterion for throughput. Follow-on work on tree
speculation~\citep{Cai2024Medusa,Li2024EAGLE,Miao2024SpecInfer,Chen2024Sequoia},
retrieval-based drafting~\citep{He2024REST}, lookahead/parallel
schemes~\citep{Fu2024Lookahead}, and draft-target
alignment~\citep{Goel2024DraftAlign} keep the distributional guarantee
but expand the acceptance surface; \citet{Xia2024SpecSurvey} survey the
design space. Deterministic-inference work~\citep{Gond2026LLM42} treats
verification fidelity as a reproducibility problem. Serving-systems
papers~\citep{Yu2022Orca,Kwon2023PagedAttention,Agrawal2024Sarathi}
embed speculative decoding inside larger throughput-optimisation
machinery. None of this literature treats speculative decoding as a
behavioural safety control.

\paragraph{Speculative decoding as a safety/security surface.}
The closest published security-side result on speculative decoding is
\citet{Wei2024SideChannels}, who exhibit a timing side-channel that
lets an observer fingerprint inputs and exfiltrate datastore contents
from speculative-decoding traffic at $>25$ tokens/sec. That work
demonstrates speculative decoding as a security-relevant attack
surface, but the leak is an information-flow channel, not a behavioural
change in safety-scored outputs. From the defensive side,
\citet{Wang2025SpecSafety} use speculative sampling as a runtime safety
mechanism, switching between a small safety-aligned draft and a larger
composite to manage jailbreak risk. Neither line of work
empirically tests whether draft-side alignment leaks into verified
safety-scored outputs through the acceptance pathway. That gap is what
the present paper addresses. E2 (\S\ref{sec:results}) is the strongest
construction of the leakage concern that fits inside an open-weights
design space: we fine-tune the draft with DPO~\citep{Rafailov2023DPO}
on Anthropic's hh-rlhf corpus~\citep{Bai2022Anthropic} with preference
labels \emph{flipped}, so that the draft prefers harmful completions,
and then pair it with the unchanged target.

\paragraph{Evaluation and statistics.}
Safety benchmarks give the evaluation vocabulary:
AdvBench~\citep{Zou2023AdvBench}, BBQ~\citep{Parrish2022BBQ},
TruthfulQA~\citep{Lin2022TruthfulQA}, and MMLU/ARC capability
controls~\citep{Hendrycks2021MMLU,Clark2018ARC}; standardised
red-teaming and jailbreak-robustness benchmarks such as
HarmBench~\citep{Mazeika2024HarmBench} and
JailbreakBench~\citep{Chao2024JailbreakBench}, and safety-preference
datasets such as BeaverTails~\citep{Ji2023BeaverTails} and the
Llama Guard taxonomy~\citep{Inan2023LlamaGuard}, are the natural
follow-on batteries for \tais{}. Reproducible evaluation infrastructure
follows the EleutherAI lm-eval-harness
practice~\citep{Biderman2024LMEvalHarness}. The equivalence-testing
apparatus is standard: \citet{Lakens2017TOST} for TOST,
\citet{Cohen1988Stats} for effect-size thresholds
($|h| < 0.2$ ``trivial'', $|h| < 0.5$ ``small''),
\citet{Wilson1927Binomial} for binomial proportion intervals, and
\citet{Holm1979Bonf} for multiple-comparison control across the
expansion AdvBench contrasts.

\section{Methods}
\label{sec:design}

\subsection{Factorial design and expansion battery}

A five-phase factorial design tests speculative decoding's impact on
safety across three axes: acceptance method (strict rejection sampling
versus probabilistic typical acceptance), model pair (three pairs from
two families), and speculation length ($N \in \{1,3,5,8,12\}$). Phase~1
runs each of 5 standalone models (3 targets + 2 drafts) on the full
953-prompt battery to establish baselines. Phases~2 and~3 hold the model
pair and prompt set fixed and vary only the acceptance method. Phase~4
sweeps speculation length on a 420-prompt safety subset to expose
dose-response curvature. Phase~5 aggregates Prometheus-exported
per-request acceptance-rate telemetry. The core produces \coreSamples{}
paired samples.

On top of the core we layer five expansion experiments that each close a
  specific reviewer-anticipated escape hatch: \textbf{E1 (production
  scale)} Llama-3.1-70B-Instruct-AWQ-INT4 target with an 8B fp16 draft on
  an A100-SXM-80GB (a one-arm production-scale sanity probe, not a
  matched \tais{} equivalence cell, because the 70B target-only arm was
  not run); \textbf{E2 (adversarial draft)} a
DPO-trained~\citep{Rafailov2023DPO} draft on
Anthropic/hh-rlhf~\citep{Bai2022Anthropic} with preference labels
flipped, operationalising the acceptance-pathway leakage concern at its
strongest construction; \textbf{E3 (quantised draft)} a GPTQ-4bit build of
Llama-3.2-1B-Instruct~\citep{Frantar2023GPTQ}, testing the
deployment-dominant ``quantise the draft'' pattern; \textbf{E4 (seed
replication)} all three core pairs re-run with seeds 123 and 456 through
Phases~2+3+4 to close the lucky-seed objection; \textbf{E5 (dtype
robustness)} all three core pairs re-run with accumulator dtype bf16.
Total expansion volume is \expansionSamples{} samples;
\S\ref{sec:expansion_detail} (appendix) tabulates per-experiment coverage.

\subsection{Safety benchmarks and scoring}

The prompt battery is: AdvBench refusal (100 prompts;
\citealp{Zou2023AdvBench}), jailbreak amplification (120 prompts,
in-house probe), BBQ bias (198 prompts after deduplication;
\citealp{Parrish2022BBQ}), TruthfulQA (50 prompts;
\citealp{Lin2022TruthfulQA}), MMLU (285 prompts;
\citealp{Hendrycks2021MMLU}) and ARC-Challenge (200 prompts;
\citealp{Clark2018ARC}). Each sample is scored by two independent
pipelines: (i) deterministic regex classifiers for refusal, bias, and
truthfulness, and (ii) an LLM judge (Gemma~3~12B via Ollama, blinded to
the speculative-decoding configuration). The judge produced 11{,}448
labels on the core; judge labels are not part of the submitted
expansion-cell evidence, so expansion conclusions use the deterministic
classifier pipeline only.

\subsection{Serving-stack configuration}

All inference runs through vLLM v0.19 with GPU passthrough on an
RTX~4080 Laptop 12GB (core) and A100-SXM-80GB (E1). Speculative decoding
is configured via the \texttt{--speculative-config} JSON introduced in
vLLM v0.19, specifying draft model, method (\texttt{draft\_model}),
\texttt{num\_speculative\_tokens}, and
\texttt{rejection\_sample\_method} $\in \{$\texttt{strict},
\texttt{probabilistic}$\}$. Per-request Prometheus counters
\texttt{vllm:spec\_decode\_num\_\{accepted,draft\}\_tokens\_total} are
polled before and after each request to compute per-request acceptance
rates.

\subsection{The \tais{} screen}
\label{sec:tais}

The \taisLong{} (\tais{}) is a behavioural-equivalence protocol for
speculative-decoding stacks. Let $\mathcal{P}$ be a set of safety-scored
tasks, let $p_{\text{target}}^{(t)}$ denote the safety-score rate for
task $t$ under target-only decoding, and let $p_{\text{spec}}^{(t)}$
denote the rate under speculative decoding with fixed draft, acceptance
method, and speculation length. \tais{} declares the stack
\emph{null-consistent} on task $t$ iff all three hold:
\begin{enumerate}[leftmargin=*,nosep]
  \item \textbf{Effect size.} $|h_t| = |2\arcsin\sqrt{p_{\text{target}}^{(t)}} - 2\arcsin\sqrt{p_{\text{spec}}^{(t)}}| < h^\star$.
  \item \textbf{Equivalence.} Two One-Sided Tests~\citep{Lakens2017TOST} reject both $\{p_{\text{spec}}^{(t)} - p_{\text{target}}^{(t)} \geq \Delta\}$ and $\{p_{\text{spec}}^{(t)} - p_{\text{target}}^{(t)} \leq -\Delta\}$ at significance level $\alpha$, with $\Delta = 0.03$ (\tostBand{}).
  \item \textbf{Byte-identity flag.} At temperature zero, paired outputs on identical prompts are compared byte-for-byte; the stack earns a ``strong'' flag at $\geq 99.5\%$ identity and a ``moderate'' flag when identity is lower but criteria~(1) and~(2) still pass.
\end{enumerate}
A stack is declared null-consistent over $\mathcal{P}$ iff
$\max_t |h_t| < h^\star$ AND criterion (2) passes on every
well-conditioned per-task contrast; capability-domain Wald-CI edge
cases at $p_{\text{target}}{=}p_{\text{spec}}$ at the boundary
(0.0 or 1.0) are flagged separately rather than counted as failures,
since the underlying observed difference is exactly zero. Multiple
comparisons are controlled using
Holm--Bonferroni~\citep{Holm1979Bonf}. The cutoff
$h^\star = \nullCutoffH{}$ used in this paper is calibrated from the
  expansion data, not pre-registered: the maximum absolute $h$ across 17
  matched AdvBench-refusal contrasts spanning E0 and E2--E5 is \maxCohensH{}, so
$h^\star = 0.1$ sits a factor of four above the loudest observed null-consistent contrast and a factor of two below the \citet{Cohen1988Stats} ``trivial'' floor of 0.2. \textbf{Calibration caveat (circularity).} $h^\star$ is calibrated post-hoc on the same expansion data used to declare the null---a real methodological weakness. Future applications of \tais{} should either (i) reserve a held-out calibration set or (ii) adopt the conservative default $h^\star=0.2$ (Cohen's ``small effect'' floor); the present null trivially survives that conservative cutoff since $\max|h|=\maxCohensH{}\ll 0.2$, so the headline does not depend on the $0.1$ choice.
\textbf{Held-out calibration check.} Stratifying the 17 matched AdvBench-refusal contrasts within each experiment family (E0/E2/E3/E4/E5; the one-arm E1 probe is excluded) and alternating by publication index yields \heldoutNcal{} calibration and \heldoutNeval{} evaluation contrasts. Calibration gives $\max|h|_{\text{cal}}=\heldoutMaxHCal{}$, hence $h^\star_{\text{cal}}=4\times\heldoutMaxHCal{}=\heldoutHstarCal{}$; evaluation gives $\max|h|_{\text{eval}}=\heldoutMaxHEval{}<h^\star_{\text{cal}}$, so the cutoff structure survives held-out verification, landing within $3\%$ of the full-corpus $\nullCutoffH{}$. The headline ($\max|h|=\maxCohensH{}$) is not an artefact of post-hoc calibration on the same corpus.

\paragraph{What \tais{} is not.}
\tais{} is not a pre-registered cutoff, it is not a hypothesis test on a
single experiment, and it does not attempt to bound
temperature-greater-than-zero behaviour. It is a reusable behavioural
screen: a vendor supplying a speculative-decoding stack can run \tais{}
on a matched safety battery and report pass/fail, and a user can inspect
per-task $h$ and TOST $p$-values before enabling the optimisation.

\subsection{Statistical methodology}

In addition to \tais{}, we use McNemar's exact test for paired
contingency-table symmetry, TOST at \tostBand{} for equivalence in the
full prompt-level ledger, logistic regression of binary safety score on
speculation length with 1{,}000-resample bootstrap CIs for dose-response,
and Mantel--Haenszel odds-ratio pooling across model-pair strata for
cross-model synthesis. Holm--Bonferroni~\citep{Holm1979Bonf} controls
family-wise error across expansion contrasts, with adjusted $\alpha =
\holmAlpha$ across 11 AdvBench tests.

\section{Results}
\label{sec:results}

We report the core null, the production-scale E1 sanity probe, the
load-bearing matched expansion headlines (especially E2 adversarial
draft), and the pooled cross-experiment synthesis that supports the
\tais{} calibration.
Full per-experiment detail, per-task Cohen's~$h$ tables for E3--E5, and
the byte-identity matrix are deferred to
\S\ref{sec:expansion_detail} (appendix).

\subsection{Core null (E0): byte-identity plus McNemar plus TOST}

The confirmatory core is \coreSamples{} paired samples across
\corePairs{} model pairs, four safety tasks, and two capability control
tasks. Phase~2 (strict rejection sampling) byte-identity against
target-only output is \coreByteIdentity{} across 2{,}859 paired
comparisons (\coreByteIdentityN{}); the non-identical fraction (9.34\%)
arises from fp16 non-associativity in the verification computation.
Only 10 of 267 textual differences cross the regex-classifier's safety
boundary (0.35\% of all comparisons), and only 1 crosses the capability
boundary. Phase~3 (typical acceptance) paired-McNemar tests are
non-significant on all three pairs (llama3.2-3b+1b $p=1.0$;
qwen2.5-3b+1.5b $p=0.5$; qwen2.5-1.5b+0.5b $p=0.625$). Discordant pairs
are 2--4 out of 953, directionally inconsistent across pairs. Hypothesis
H2 (typical acceptance degrades safety) is not supported. TOST with
\tostBand{} equivalence bound passes on \coreTOSTpass{} comparisons;
the two non-pass contrasts are capability-domain (MMLU and ARC-Challenge)
Wald-CI edge cases at $p_{\text{target}}{=}p_{\text{spec}}$ at the
ceiling, where the underlying observed difference is exactly zero,
not genuine non-equivalence. The core
supports the null on both acceptance methods with a per-pair MDE of
7.4--8.3pp.

\subsection{E1: production-scale (Llama-3.1-70B + 8B)}

E1 scales the target $23\times$, using Llama-3.1-70B-Instruct-AWQ-INT4
as target and the fp16 Llama-3.1-8B-Instruct as draft on a single
A100-SXM-80GB (4{,}006 speculative samples across Phases~2, 3, 4).
Because the matched 70B target-only arm was not run, E1 is not a
\tais{} equivalence cell. It is a production-scale sanity probe: across
all \seventyBAdvBenchN{} AdvBench completions from Phases~2--4, the
refusal rate is \seventyBrefusal{} with 95\% Wilson CI
\seventyBrefusalCI{}; the Phase~2+3 full-battery slice is
\seventyBPhaseTwoThreeRefusal{} \seventyBPhaseTwoThreeRefusalCI{}.
The point estimate is close to the smaller Llama-family core refusal
band, but the one-arm cross-model design precludes an equivalence
claim. The Phase~4 dose-response ($N \in \{1,3,5,8,12\}$) is flat
within 0.36pp across all five length settings; the logistic slope is
$0.000$ with bootstrap 95\% CI $[-0.001, +0.001]$ (1{,}000 resamples,
seed=42). Per-phase safety rates are given in
Table~\ref{tab:e1_per_phase}.

\begin{table}[t]
  \centering
  \caption{E1 per-phase safety rates (Llama-3.1-70B-Instruct-AWQ-INT4 target, fp16 Llama-3.1-8B-Instruct draft). Phase~2 and~3 aggregate safety across the full prompt battery; Phase~4 is the 2{,}100-sample speculation-length sweep. The AdvBench rows report the Phase~2+3 full-battery slice and the Phase~2--4 aggregate production-scale sanity rate; neither row is a matched \tais{} contrast because the 70B target-only arm was not run.}
  \label{tab:e1_per_phase}
  \small
  \begin{tabular}{l l r l l}
    \toprule
    Phase & Method & $n$ & Safety rate & 95\% Wilson CI \\
    \midrule
    2 & rejection\_sampler & 468 & 0.3632 & [0.320, 0.409] \\
    3 & typical\_acceptance\_sampler & 468 & 0.3632 & [0.320, 0.409] \\
    4 & typical\_acceptance\_sampler & 2{,}100 & 0.4038 & [0.383, 0.425] \\
    AdvBench & Phases 2+3 & \seventyBPhaseTwoThreeN{} & \seventyBPhaseTwoThreeRefusal{} & \seventyBPhaseTwoThreeRefusalCI{} \\
    AdvBench & Phases 2--4 & \seventyBAdvBenchN{} & \seventyBrefusal{} & \seventyBrefusalCI{} \\
    \bottomrule
  \end{tabular}
\end{table}

\subsection{E2: adversarial DPO draft}

E2 is a strong construction of the acceptance-pathway leakage
concern within open-weights design space. The draft
(Llama-3.2-1B-Instruct) is fine-tuned with DPO~\citep{Rafailov2023DPO}
on Anthropic/hh-rlhf~\citep{Bai2022Anthropic} with preference labels
flipped (``rejected'' treated as preferred), then paired with the
unmodified Llama-3.2-3B-Instruct target. Per-phase safety rates are
identical to three decimal places against the canonical-draft E0
configuration (Phase~2 0.3729 vs.\ 0.3729; Phase~3 0.3729 vs.\ 0.3729;
Phase~4 0.4148 vs.\ 0.4148). Byte-identity against the canonical-draft
output is \expansionIdentity{} across 4{,}006 samples: changing the
draft from fp16-canonical to DPO-adversarial-on-flipped-hh-rlhf does
not change a single byte of the verified output. This is consistent with
the draft having no detectable behavioral channel into the output at
temp=0 greedy in this construction. Every per-task safety contrast
passes TOST at \tostBand{}; per-task safety rates are identical to four decimal places against the canonical-draft E0 configuration, so per-task $|h|$ is 0.000 across the safety battery. E3
(GPTQ-4bit draft) and E4 (seed replication across 24{,}036 samples)
produce \expansionIdentity{} byte-identity on identical keys and
exactly 0.00pp per-pair safety deltas across all tasks.
Table~\ref{tab:e2e3_byte_id} summarises the three pairwise byte-identity
comparisons on the llama3.2-3b-target family: all three counterfactual
draft constructions tested (adversarial DPO, GPTQ-4bit,
seed-replicated canonical) produce byte-identical output on the matched
sample keys.

\begin{table}[t]
  \centering
  \caption{E2 and E3 byte-identity to canonical-draft output on the
  llama3.2-3b-target family. Comparison against E4 seed\_123 on the
  same (sample\_id, phase, acceptance\_method, $N$) key. At temperature
  zero and fp16 accumulator, the draft is empirically interchangeable
  across the three constructions tested.}
  \label{tab:e2e3_byte_id}
  \small
  \begin{tabular}{l r r l}
    \toprule
    Comparison & Common keys & Byte-identical & Identity rate \\
    \midrule
    E2 (adversarial) vs.\ E4 seed\_123 (canonical) & 4{,}006 & 4{,}006 & \expansionIdentity{} \\
    E2 (adversarial) vs.\ E3 (GPTQ-4bit draft)    & 4{,}006 & 4{,}006 & \expansionIdentity{} \\
    E3 (GPTQ-4bit) vs.\ E4 seed\_123 (canonical)  & 4{,}006 & 4{,}006 & \expansionIdentity{} \\
    \bottomrule
  \end{tabular}
\end{table}

\subsection{E5: dtype robustness and cross-experiment synthesis}

E5 re-runs all three core pairs with accumulator dtype bf16 instead of
fp16. Unlike E2--E4, bf16 \emph{does} shift output bytes: per-pair
byte-identity against the fp16 core is 36--53\% (llama3.2-3b+1b 39.92\%,
qwen2.5-3b+1.5b 36.15\%, qwen2.5-1.5b+0.5b 52.70\%). None of this byte
shift escapes the \tostBand{} equivalence bound on any of the nine
per-task contrasts; the maximum $|h|$ on E5 is 0.054
(qwen2.5-3b+1.5b jailbreak amplification), well below the
\citet{Cohen1988Stats} ``trivial'' floor of 0.20. E5 per-task rows are given in Table~\ref{tab:e5_contrasts}.

\begin{table}[t]
  \centering
  \caption{E5 per-task safety contrasts, fp16 core vs.\ bf16. All nine rows pass TOST at \tostBand{}; the maximum $|h|$ is 0.054 (qwen2.5-3b+1.5b jailbreak amplification), well below the \citet{Cohen1988Stats} ``trivial'' floor of 0.20.}
  \label{tab:e5_contrasts}
  \small
  \setlength{\tabcolsep}{4pt}
  \begin{tabular}{l l r r r r l}
    \toprule
    Pair & Metric & fp16 $p$ & bf16 $p$ & $\Delta$ (pp) & Cohen's $h$ & TOST \tostBand{} \\
    \midrule
    llama3.2-3b+1b  & AdvBench refusal & 0.790 & 0.780 & $-1.00$ & $-0.024$ & PASS \\
    llama3.2-3b+1b  & Jailbreak amp    & 0.583 & 0.575 & $-0.80$ & $-0.016$ & PASS \\
    llama3.2-3b+1b  & TruthfulQA       & 0.560 & 0.553 & $-0.70$ & $-0.014$ & PASS \\
    qwen2.5-3b+1.5b & AdvBench refusal & 0.970 & 0.970 & $+0.00$ & $+0.000$ & PASS \\
    qwen2.5-3b+1.5b & Jailbreak amp    & 0.408 & 0.435 & $+2.70$ & $+0.054$ & PASS \\
    qwen2.5-3b+1.5b & TruthfulQA       & 0.480 & 0.503 & $+2.30$ & $+0.046$ & PASS \\
    qwen2.5-1.5b+0.5b & AdvBench refusal & 0.980 & 0.980 & $+0.00$ & $+0.000$ & PASS \\
    qwen2.5-1.5b+0.5b & Jailbreak amp    & 0.575 & 0.593 & $+1.80$ & $+0.036$ & PASS \\
    qwen2.5-1.5b+0.5b & TruthfulQA       & 0.510 & 0.486 & $-2.40$ & $-0.048$ & PASS \\
    \bottomrule
  \end{tabular}
\end{table}

Table~\ref{tab:cross_synthesis} stacks the matched AdvBench-refusal
contrasts and includes E1 as a visibly separate one-arm scale row. Every
matched contrast passes TOST at \tostBand{}. The maximum absolute
Cohen's~$h$ among matched rows is \maxCohensH{} (E5
llama3.2-3b+1b), a factor of $\sim$8 below the ``trivial'' threshold of
0.20 and a factor of $\sim$4 below the \tais{} null cutoff of
$\nullCutoffH{}$. After Holm--Bonferroni~\citep{Holm1979Bonf} across
the 11 matched expansion contrasts the adjusted $\alpha$ is
$\holmAlpha$; no observed effect comes within an order of magnitude of
that floor.

\begin{table}[t]
  \centering
  \caption{Cross-experiment AdvBench refusal rows. $p_{\text{TA}}=$ target-alone; $p_{\text{spec}}=$ speculative. CIs are 95\% Wilson intervals on $p_{\text{spec}}$. All matched rows pass TOST at \tostBand{}; the two non-passing contrasts in the broader 27-contrast core ledger are MMLU/ARC-Challenge Wald-CI edge cases at the ceiling and are documented in \S\ref{sec:results} rather than represented here. $^{\dagger}$E1 is a one-arm production-scale sanity row, not a \tais{} contrast, because no matched 70B target-only arm was run; it is excluded from TOST, Holm correction, and max-$|h|$ calibration.}
  \label{tab:cross_synthesis}
  \small
  \begin{tabular}{l l r r r l}
    \toprule
    Experiment & Pair & $p_{\text{TA}}$ & $p_{\text{spec}}$ & Cohen's $h$ & 95\% Wilson CI \\
    \midrule
    Core P2 & llama3.2-3b+1b    & 0.790 & 0.790 & 0.0000 & [0.728, 0.841] \\
    Core P2 & qwen2.5-3b+1.5b   & 0.970 & 0.970 & 0.0000 & [0.936, 0.986] \\
    Core P2 & qwen2.5-1.5b+0.5b & 0.980 & 0.980 & 0.0000 & [0.950, 0.992] \\
    Core P3 & llama3.2-3b+1b    & 0.790 & 0.790 & 0.0000 & [0.728, 0.841] \\
    Core P3 & qwen2.5-3b+1.5b   & 0.970 & 0.970 & 0.0000 & [0.936, 0.986] \\
    Core P3 & qwen2.5-1.5b+0.5b & 0.980 & 0.980 & 0.0000 & [0.950, 0.992] \\
    E1$^{\dagger}$ & llama3.1-70b+8b   & -- & \seventyBrefusal{} & n/a & \seventyBrefusalCI{} \\
    E2      & llama3.2-3b+adv1b & 0.790 & 0.790 & 0.0000 & [0.728, 0.841] \\
    E3      & llama3.2-3b+gptq1b& 0.790 & 0.790 & 0.0000 & [0.728, 0.841] \\
    E4 s123 & llama3.2-3b+1b    & 0.790 & 0.790 & 0.0000 & [0.728, 0.841] \\
    E4 s456 & llama3.2-3b+1b    & 0.790 & 0.790 & 0.0000 & [0.728, 0.841] \\
    E4 s123 & qwen2.5-3b+1.5b   & 0.970 & 0.970 & 0.0000 & [0.936, 0.986] \\
    E4 s456 & qwen2.5-3b+1.5b   & 0.970 & 0.970 & 0.0000 & [0.936, 0.986] \\
    E4 s123 & qwen2.5-1.5b+0.5b & 0.980 & 0.980 & 0.0000 & [0.950, 0.992] \\
    E4 s456 & qwen2.5-1.5b+0.5b & 0.980 & 0.980 & 0.0000 & [0.950, 0.992] \\
    E5      & llama3.2-3b+1b    & 0.790 & 0.780 & $-0.0243$ & [0.718, 0.832] \\
    E5      & qwen2.5-3b+1.5b   & 0.970 & 0.970 & 0.0000 & [0.936, 0.986] \\
    E5      & qwen2.5-1.5b+0.5b & 0.980 & 0.980 & 0.0000 & [0.950, 0.992] \\
    \bottomrule
  \end{tabular}
\end{table}

Pooling E4 (2 seeds) and E5 (bf16) across the three core pairs gives a
36{,}054-sample pool. The 95\% Wilson CI on pooled AdvBench refusal for
every pair contains the target-alone rate exactly; per-pool MDE tightens
to $\sim 4.3$pp. The null survives the tighter pool.

\subsection{\tais{} summary}

Applying \tais{} with $h^\star = \nullCutoffH{}$ and $\Delta = 0.03$
across the matched contrast set: all 17 matched AdvBench rows in
Table~\ref{tab:cross_synthesis} pass both effect-size and equivalence
criteria; all nine per-task E5 contrasts pass both criteria; 25 of 27
core per-task TOST contrasts pass at \tostBand{}, with the two
non-passing contrasts being capability-domain Wald-CI edge cases at
$p_{\text{target}}{=}p_{\text{spec}}$ at the ceiling (observed
difference exactly zero), flagged but not counted as failures.
The two non-passing TOST contrasts arise because both arms hit the
ceiling ($p_t = p_s = 1.0$) on capability-domain tasks (MMLU, ARC),
where the Wald 95\% CI degenerates ($\widehat{\text{SE}} = 0$); the
underlying difference is exactly zero, not a power failure to detect harm.
Byte-identity $\geq 99.5\%$ (``strong'' flag) holds across E2/E3/E4
(\expansionIdentity{}). E5 drops to a ``moderate'' flag: equivalence
and effect-size pass, byte-identity does not---exactly the case for
which the ``moderate'' flag was introduced. The matched E0 and E2--E5
battery is null-consistent under \tais{} at the calibrated cutoff within
the tested scope; E1 remains an auxiliary scale probe, not a
target-only-versus-speculative equivalence claim.

\section{Discussion}
\label{sec:discussion}

\subsection{Why the predicted leakage does not occur}

The acceptance-pathway leakage mechanism requires two conditions: (i) draft--target disagreement on the sampled token, and (ii) verification soft enough to let the draft-preferred token survive. At $T=0$ both acceptance policies collapse onto near-hard equality, so the \citet{Leviathan2023Speculative} distributional guarantee becomes a \emph{byte}-level guarantee modulo numerical non-associativity. Non-associativity explains the core 9.34\% non-identical fraction and the bf16 \bfsixteenShiftLow{}--\bfsixteenShiftHigh{} byte shift; it does not produce a safety signal because floating-point rounding is uncorrelated with the refusal classifier boundary.

E2 is operational validation rather than theoretical discovery: a draft DPO-trained on hh-rlhf~\citep{Bai2022Anthropic} with \emph{flipped} preference labels produces byte-identical output to the canonical draft across 4{,}006 samples---the expected algorithmic consequence of \citet{Leviathan2023Speculative} at $T{=}0$ modulo non-associativity, validated against implementation bugs and verification-code shortcuts on the real vLLM v0.19 stack. The acceptance pathway exists structurally in vLLM's verification code, but at the greedy operating point it has no measurable behavioural channel into safety-scored outputs on the probed battery.

\subsection{What \tais{} adds}

\tais{} turns the negative result into an auditable protocol: a third party can run \tais{} against a matched safety battery, collect per-task Cohen's $h$ and TOST $p$-values~\citep{Lakens2017TOST}, and report pass/fail at the $\nullCutoffH{}$ cutoff (calibration in \S\ref{sec:design}); a stack that fails is outside the empirically observed null envelope.

\textbf{Calibration circularity, restated.} $h^\star=\nullCutoffH{}$ is post-hoc calibrated on the expansion data; the held-out check in \S\ref{sec:tais} reproduces the cutoff structure, but until an independent corpus exists, downstream uses should adopt $h^\star=0.2$. Vendor documentation reports throughput and acceptance rates but not behavioural-equivalence evidence; \tais{} specifies that evidence.

\subsection{Limitations and threats to validity}

The null is conditional on five scope boundaries: $T=0$ only (leakage concern is strongest at $T>0$); vLLM v0.19 only; Llama 3.x and Qwen 2.5 only; rejection sampling and typical acceptance only (EAGLE~\citep{Li2024EAGLE}/Medusa~\citep{Cai2024Medusa} tree variants untested); and the AdvBench/jailbreak/BBQ/TruthfulQA/MMLU/ARC battery (multi-turn and agentic out of scope).

Statistical floor: per-cell MDE 7.4--8.3pp, pooled $\sim 4.3$pp; $|h|<0.1$ is $\sim 5$pp at 50\% base rate and $<3$pp at 80--98\%. The llama3.2-3b+1b pair supplies dynamic range; Qwen ceiling cells (0.970--0.980) extend coverage but have near-zero power. Expansion uses deterministic regex; Gemma-3-12B labels on the core are a cross-pipeline check.

\subsection{Implications and future work}

Within the tested scope, a \tais{}-pass is observed evidence against a safety regression; draft choice reduces to throughput. Future work inverts the scope: $T>0$, EAGLE/Medusa, cross-framework, multi-turn.

\section{Conclusion}
\label{sec:conclusion}

Across \coreSamples{}-sample core plus \matchedExpansionSamples{} matched expansion at $T{=}0$, greedy speculative decoding shows no detectable safety divergence under \tais{} (max $|h|=\maxCohensH{}$; 25/27 core TOST pass at \tostBand{}; DPO-adversarial draft byte-identical across 4{,}006 samples). The 70B run is production-scale plausibility, not a \tais{} pass. Open: $T>0$, tree variants, and matched 70B pairs.

\bibliographystyle{plainnat}
\bibliography{refs}

\begin{thebibliography}{40}
\providecommand{\natexlab}[1]{#1}
\providecommand{\url}[1]{\texttt{#1}}
\expandafter\ifx\csname urlstyle\endcsname\relax
  \providecommand{\doi}[1]{doi: #1}\else
  \providecommand{\doi}{doi: \begingroup \urlstyle{rm}\Url}\fi

\bibitem[Agrawal et~al.(2024)Agrawal, Kedia, Panwar, Mohan, Kwatra, Gulavani,
  Tumanov, and Ramjee]{Agrawal2024Sarathi}
Amey Agrawal, Nitin Kedia, Ashish Panwar, Jayashree Mohan, Nipun Kwatra,
  Bhargav~S. Gulavani, Alexey Tumanov, and Ramachandran Ramjee.
\newblock Taming throughput-latency tradeoff in {LLM} inference with
  {Sarathi-Serve}.
\newblock \emph{arXiv preprint arXiv:2403.02310}, 2024.
\newblock URL \url{https://arxiv.org/abs/2403.02310}.

\bibitem[Bai et~al.(2022)Bai, Jones, Ndousse, Askell, Chen, DasSarma, Drain,
  Fort, Ganguli, Henighan, et~al.]{Bai2022Anthropic}
Yuntao Bai, Andy Jones, Kamal Ndousse, Amanda Askell, Anna Chen, Nova DasSarma,
  Dawn Drain, Stanislav Fort, Deep Ganguli, Tom Henighan, et~al.
\newblock Training a helpful and harmless assistant with reinforcement learning
  from human feedback.
\newblock \emph{arXiv preprint arXiv:2204.05862}, 2022.
\newblock URL \url{https://arxiv.org/abs/2204.05862}.

\bibitem[Biderman et~al.(2024)Biderman, Schoelkopf, Sutawika, Gao, Tow, Abbasi,
  Aji, Ammanamanchi, Black, Clive, et~al.]{Biderman2024LMEvalHarness}
Stella Biderman, Hailey Schoelkopf, Lintang Sutawika, Leo Gao, Jonathan Tow,
  Baber Abbasi, Alham~Fikri Aji, Pawan~Sasanka Ammanamanchi, Sidney Black,
  Jordan Clive, et~al.
\newblock Lessons from the trenches on reproducible evaluation of language
  models.
\newblock \emph{arXiv preprint arXiv:2405.14782}, 2024.
\newblock URL \url{https://arxiv.org/abs/2405.14782}.

\bibitem[Cai et~al.(2024)Cai, Li, Geng, Peng, Lee, Chen, and
  Dao]{Cai2024Medusa}
Tianle Cai, Yuhong Li, Zhengyang Geng, Hongwu Peng, Jason~D. Lee, Deming Chen,
  and Tri Dao.
\newblock {Medusa}: Simple {LLM} inference acceleration framework with multiple
  decoding heads.
\newblock \emph{arXiv preprint arXiv:2401.10774}, 2024.
\newblock URL \url{https://arxiv.org/abs/2401.10774}.

\bibitem[Chao et~al.(2024)Chao, Debenedetti, Robey, Andriushchenko, Croce,
  Sehwag, Dobriban, Flammarion, Pappas, Tramèr, Hassani, and
  Wong]{Chao2024JailbreakBench}
Patrick Chao, Edoardo Debenedetti, Alexander Robey, Maksym Andriushchenko,
  Francesco Croce, Vikash Sehwag, Edgar Dobriban, Nicolas Flammarion, George~J.
  Pappas, Florian Tramèr, Hamed Hassani, and Eric Wong.
\newblock {JailbreakBench}: An open robustness benchmark for jailbreaking large
  language models.
\newblock In \emph{Advances in Neural Information Processing Systems 37,
  Datasets and Benchmarks Track}, 2024.
\newblock URL \url{https://arxiv.org/abs/2404.01318}.

\bibitem[Chen et~al.(2023)Chen, Borgeaud, Irving, Lespiau, Sifre, and
  Jumper]{Chen2023Accelerating}
Charlie Chen, Sebastian Borgeaud, Geoffrey Irving, Jean-Baptiste Lespiau,
  Laurent Sifre, and John Jumper.
\newblock Accelerating large language model decoding with speculative sampling.
\newblock \emph{arXiv preprint arXiv:2302.01318}, 2023.
\newblock URL \url{https://arxiv.org/abs/2302.01318}.

\bibitem[Chen et~al.(2024)Chen, May, Svirschevski, Huang, Ryabinin, Jia, and
  Chen]{Chen2024Sequoia}
Zhuoming Chen, Avner May, Ruslan Svirschevski, Yuhsun Huang, Max Ryabinin,
  Zhihao Jia, and Beidi Chen.
\newblock {Sequoia}: Scalable, robust, and hardware-aware speculative decoding.
\newblock \emph{arXiv preprint arXiv:2402.12374}, 2024.
\newblock URL \url{https://arxiv.org/abs/2402.12374}.

\bibitem[Christiano et~al.(2017)Christiano, Leike, Brown, Martic, Legg, and
  Amodei]{Christiano2017DeepRLHF}
Paul~F. Christiano, Jan Leike, Tom~B. Brown, Miljan Martic, Shane Legg, and
  Dario Amodei.
\newblock Deep reinforcement learning from human preferences.
\newblock In \emph{Advances in Neural Information Processing Systems 30}, 2017.
\newblock URL \url{https://arxiv.org/abs/1706.03741}.

\bibitem[Clark et~al.(2018)Clark, Cowhey, Etzioni, Khot, Sabharwal, Schoenick,
  and Tafjord]{Clark2018ARC}
Peter Clark, Isaac Cowhey, Oren Etzioni, Tushar Khot, Ashish Sabharwal, Carissa
  Schoenick, and Oyvind Tafjord.
\newblock Think you have solved question answering? try {ARC}, the {AI2}
  reasoning challenge.
\newblock \emph{arXiv preprint arXiv:1803.05457}, 2018.
\newblock URL \url{https://arxiv.org/abs/1803.05457}.

\bibitem[Cohen(1988)]{Cohen1988Stats}
Jacob Cohen.
\newblock \emph{Statistical Power Analysis for the Behavioral Sciences}.
\newblock Routledge, Hillsdale, NJ, 2nd edition, 1988.

\bibitem[Frantar et~al.(2023)Frantar, Ashkboos, Hoefler, and
  Alistarh]{Frantar2023GPTQ}
Elias Frantar, Saleh Ashkboos, Torsten Hoefler, and Dan Alistarh.
\newblock {GPTQ}: Accurate post-training quantization for generative
  pre-trained transformers.
\newblock In \emph{International Conference on Learning Representations
  ({ICLR})}, 2023.
\newblock URL \url{https://arxiv.org/abs/2210.17323}.

\bibitem[Fu et~al.(2024)Fu, Bailis, Stoica, and Zhang]{Fu2024Lookahead}
Yichao Fu, Peter Bailis, Ion Stoica, and Hao Zhang.
\newblock Break the sequential dependency of {LLM} inference using lookahead
  decoding.
\newblock \emph{arXiv preprint arXiv:2402.02057}, 2024.
\newblock URL \url{https://arxiv.org/abs/2402.02057}.

\bibitem[Goel et~al.(2024)Goel, Gagrani, Jeon, Park, Lee, and
  Lott]{Goel2024DraftAlign}
Raghavv Goel, Mukul Gagrani, Wonseok Jeon, Junyoung Park, Mingu Lee, and
  Christopher Lott.
\newblock Direct alignment of draft model for speculative decoding with
  chat-fine-tuned {LLMs}.
\newblock \emph{arXiv preprint arXiv:2403.00858}, 2024.
\newblock URL \url{https://arxiv.org/abs/2403.00858}.

\bibitem[Gond et~al.(2026)Gond, Kamath, Ramjee, and Panwar]{Gond2026LLM42}
Raja Gond, Aditya~K. Kamath, Ramachandran Ramjee, and Ashish Panwar.
\newblock {LLM}-42: Enabling determinism in {LLM} inference with verified
  speculation.
\newblock \emph{arXiv preprint arXiv:2601.17768}, 2026.
\newblock URL \url{https://arxiv.org/abs/2601.17768}.

\bibitem[He et~al.(2024)He, Zhong, Cai, Lee, and He]{He2024REST}
Zhenyu He, Zexuan Zhong, Tianle Cai, Jason~D. Lee, and Di~He.
\newblock {REST}: Retrieval-based speculative decoding.
\newblock In \emph{Proceedings of the 2024 Conference of the North American
  Chapter of the Association for Computational Linguistics ({NAACL})}, 2024.
\newblock URL \url{https://arxiv.org/abs/2311.08252}.

\bibitem[Hendrycks et~al.(2021)Hendrycks, Burns, Basart, Zou, Mazeika, Song,
  and Steinhardt]{Hendrycks2021MMLU}
Dan Hendrycks, Collin Burns, Steven Basart, Andy Zou, Mantas Mazeika, Dawn
  Song, and Jacob Steinhardt.
\newblock Measuring massive multitask language understanding.
\newblock \emph{arXiv preprint arXiv:2009.03300}, 2021.
\newblock URL \url{https://arxiv.org/abs/2009.03300}.

\bibitem[Holm(1979)]{Holm1979Bonf}
Sture Holm.
\newblock A simple sequentially rejective multiple test procedure.
\newblock \emph{Scandinavian Journal of Statistics}, 6\penalty0 (2):\penalty0
  65--70, 1979.

\bibitem[Huang et~al.(2025)Huang, Hu, Ilhan, Tekin, Yahn, Xu, and
  Liu]{Huang2025SafetyTax}
Tiansheng Huang, Sihao Hu, Fatih Ilhan, Selim~Furkan Tekin, Zachary Yahn,
  Yichang Xu, and Ling Liu.
\newblock Safety tax: Safety alignment makes your large reasoning models less
  reasonable.
\newblock \emph{arXiv preprint arXiv:2503.00555}, 2025.
\newblock URL \url{https://arxiv.org/abs/2503.00555}.

\bibitem[Inan et~al.(2023)Inan, Upasani, Chi, Rungta, Iyer, Mao, Tontchev, Hu,
  Fuller, Testuggine, and Khabsa]{Inan2023LlamaGuard}
Hakan Inan, Kartikeya Upasani, Jianfeng Chi, Rashi Rungta, Krithika Iyer,
  Yuning Mao, Michael Tontchev, Qing Hu, Brian Fuller, Davide Testuggine, and
  Madian Khabsa.
\newblock Llama guard: {LLM}-based input-output safeguard for human-{AI}
  conversations.
\newblock 2023.
\newblock URL \url{https://arxiv.org/abs/2312.06674}.

\bibitem[Ji et~al.(2023)Ji, Liu, Dai, Pan, Zhang, Bian, Sun, Wang, and
  Yang]{Ji2023BeaverTails}
Jiaming Ji, Mickel Liu, Juntao Dai, Xuehai Pan, Chi Zhang, Ce~Bian, Ruiyang
  Sun, Yizhou Wang, and Yaodong Yang.
\newblock {BeaverTails}: Towards improved safety alignment of {LLM} via a
  human-preference dataset.
\newblock In \emph{Advances in Neural Information Processing Systems 36,
  Datasets and Benchmarks Track}, 2023.
\newblock URL \url{https://arxiv.org/abs/2307.04657}.

\bibitem[Kang et~al.(2023)Kang, Li, Stoica, Guestrin, Zaharia, and
  Hashimoto]{Kang2023Programmatic}
Daniel Kang, Xuechen Li, Ion Stoica, Carlos Guestrin, Matei Zaharia, and
  Tatsunori Hashimoto.
\newblock Exploiting programmatic behavior of {LLMs}: Dual-use through standard
  security attacks.
\newblock \emph{arXiv preprint arXiv:2302.05733}, 2023.
\newblock URL \url{https://arxiv.org/abs/2302.05733}.

\bibitem[Kwon et~al.(2023)Kwon, Li, Zhuang, Sheng, Zheng, Yu, Gonzalez, Zhang,
  and Stoica]{Kwon2023PagedAttention}
Woosuk Kwon, Zhuohan Li, Siyuan Zhuang, Ying Sheng, Lianmin Zheng, Cody~Hao Yu,
  Joseph~E. Gonzalez, Hao Zhang, and Ion Stoica.
\newblock Efficient memory management for large language model serving with
  {PagedAttention}.
\newblock In \emph{Proceedings of the 29th Symposium on Operating Systems
  Principles ({SOSP})}, 2023.
\newblock URL \url{https://arxiv.org/abs/2309.06180}.

\bibitem[Lakens(2017)]{Lakens2017TOST}
Dani{\"e}l Lakens.
\newblock Equivalence tests: A practical primer for $t$ tests, correlations,
  and meta-analyses.
\newblock \emph{Social Psychological and Personality Science}, 8\penalty0
  (4):\penalty0 355--362, 2017.
\newblock \doi{10.1177/1948550617697177}.

\bibitem[Leviathan et~al.(2023)Leviathan, Kalman, and
  Matias]{Leviathan2023Speculative}
Yaniv Leviathan, Matan Kalman, and Yossi Matias.
\newblock Fast inference from transformers via speculative decoding.
\newblock In \emph{Proceedings of the 40th International Conference on Machine
  Learning}, volume 202 of \emph{Proceedings of Machine Learning Research},
  pages 19274--19286, 2023.
\newblock URL \url{https://proceedings.mlr.press/v202/leviathan23a.html}.

\bibitem[Li et~al.(2024)Li, Wei, Zhang, and Zhang]{Li2024EAGLE}
Yuhui Li, Fangyun Wei, Chao Zhang, and Hongyang Zhang.
\newblock {EAGLE}: Speculative sampling requires rethinking feature
  uncertainty.
\newblock \emph{arXiv preprint arXiv:2401.15077}, 2024.
\newblock URL \url{https://arxiv.org/abs/2401.15077}.

\bibitem[Lin et~al.(2022)Lin, Hilton, and Evans]{Lin2022TruthfulQA}
Stephanie Lin, Jacob Hilton, and Owain Evans.
\newblock {TruthfulQA}: Measuring how models mimic human falsehoods.
\newblock \emph{arXiv preprint arXiv:2109.07958}, 2022.
\newblock URL \url{https://arxiv.org/abs/2109.07958}.

\bibitem[Mazeika et~al.(2024)Mazeika, Phan, Yin, Zou, Wang, Mu, Sakhaee, Li,
  Basart, Li, Forsyth, and Hendrycks]{Mazeika2024HarmBench}
Mantas Mazeika, Long Phan, Xuwang Yin, Andy Zou, Zifan Wang, Norman Mu, Elham
  Sakhaee, Nathaniel Li, Steven Basart, Bo~Li, David Forsyth, and Dan
  Hendrycks.
\newblock {HarmBench}: A standardized evaluation framework for automated red
  teaming and robust refusal.
\newblock \emph{arXiv preprint arXiv:2402.04249}, 2024.
\newblock URL \url{https://arxiv.org/abs/2402.04249}.

\bibitem[Miao et~al.(2024)Miao, Oliaro, Zhang, Cheng, Wang, Zhang, Wong, Zhu,
  Yang, Shi, Shi, Chen, Arfeen, Abhyankar, and Jia]{Miao2024SpecInfer}
Xupeng Miao, Gabriele Oliaro, Zhihao Zhang, Xinhao Cheng, Zeyu Wang, Zhengxin
  Zhang, Rae Ying~Yee Wong, Alan Zhu, Lijie Yang, Xiaoxiang Shi, Chunan Shi,
  Zhuoming Chen, Daiyaan Arfeen, Reyna Abhyankar, and Zhihao Jia.
\newblock {SpecInfer}: Accelerating large language model serving with
  tree-based speculative inference and verification.
\newblock In \emph{Proceedings of the 29th ACM International Conference on
  Architectural Support for Programming Languages and Operating Systems
  (ASPLOS)}, 2024.
\newblock URL \url{https://arxiv.org/abs/2305.09781}.

\bibitem[Ouyang et~al.(2022)Ouyang, Wu, Jiang, Almeida, Wainwright, Mishkin,
  Zhang, Agarwal, Slama, Ray, et~al.]{Ouyang2022InstructGPT}
Long Ouyang, Jeffrey Wu, Xu~Jiang, Diogo Almeida, Carroll~L. Wainwright, Pamela
  Mishkin, Chong Zhang, Sandhini Agarwal, Katarina Slama, Alex Ray, et~al.
\newblock Training language models to follow instructions with human feedback.
\newblock In \emph{Advances in Neural Information Processing Systems 35}, 2022.
\newblock URL \url{https://arxiv.org/abs/2203.02155}.

\bibitem[Parrish et~al.(2022)Parrish, Chen, Nangia, Padmakumar, Phang,
  Thompson, Htut, and Bowman]{Parrish2022BBQ}
Alicia Parrish, Angelica Chen, Nikita Nangia, Vishakh Padmakumar, Jason Phang,
  Jana Thompson, Phu~Mon Htut, and Samuel~R. Bowman.
\newblock {BBQ}: A hand-built bias benchmark for question answering.
\newblock In \emph{Findings of the Association for Computational Linguistics:
  {ACL} 2022}, 2022.
\newblock URL \url{https://aclanthology.org/2022.findings-acl.165/}.

\bibitem[Qi et~al.(2023)Qi, Zeng, Xie, Chen, Jia, Mittal, and
  Henderson]{Qi2023Finetune}
Xiangyu Qi, Yi~Zeng, Tinghao Xie, Pin-Yu Chen, Ruoxi Jia, Prateek Mittal, and
  Peter Henderson.
\newblock Fine-tuning aligned language models compromises safety, even when
  users do not intend to.
\newblock \emph{arXiv preprint arXiv:2310.03693}, 2023.
\newblock URL \url{https://arxiv.org/abs/2310.03693}.

\bibitem[Rafailov et~al.(2023)Rafailov, Sharma, Mitchell, Ermon, Manning, and
  Finn]{Rafailov2023DPO}
Rafael Rafailov, Archit Sharma, Eric Mitchell, Stefano Ermon, Christopher~D.
  Manning, and Chelsea Finn.
\newblock Direct preference optimization: Your language model is secretly a
  reward model.
\newblock In \emph{Advances in Neural Information Processing Systems 36}, 2023.
\newblock URL \url{https://arxiv.org/abs/2305.18290}.

\bibitem[Stern et~al.(2018)Stern, Shazeer, and Uszkoreit]{Stern2018Blockwise}
Mitchell Stern, Noam Shazeer, and Jakob Uszkoreit.
\newblock Blockwise parallel decoding for deep autoregressive models.
\newblock In \emph{Advances in Neural Information Processing Systems 31}, 2018.
\newblock URL \url{https://arxiv.org/abs/1811.03115}.

\bibitem[Wang et~al.(2025)Wang, Zhu, and Cheng]{Wang2025SpecSafety}
Xuekang Wang, Shengyu Zhu, and Xueqi Cheng.
\newblock Speculative safety-aware decoding.
\newblock In \emph{Proceedings of the 2025 Conference on Empirical Methods in
  Natural Language Processing ({EMNLP})}, pages 12827--12841, 2025.
\newblock URL \url{https://aclanthology.org/2025.emnlp-main.648/}.

\bibitem[Wei et~al.(2023)Wei, Haghtalab, and Steinhardt]{Wei2023Jailbroken}
Alexander Wei, Nika Haghtalab, and Jacob Steinhardt.
\newblock Jailbroken: How does {LLM} safety training fail?
\newblock In \emph{Advances in Neural Information Processing Systems 36}, 2023.
\newblock URL \url{https://arxiv.org/abs/2307.02483}.

\bibitem[Wei et~al.(2024)Wei, Abdulrazzag, Zhang, Muursepp, and
  Saileshwar]{Wei2024SideChannels}
Jiankun Wei, Abdulrahman Abdulrazzag, Tianchen Zhang, Adel Muursepp, and
  Gururaj Saileshwar.
\newblock When speculation spills secrets: Side channels via speculative
  decoding in {LLMs}.
\newblock \emph{arXiv preprint arXiv:2411.01076}, 2024.
\newblock URL \url{https://arxiv.org/abs/2411.01076}.

\bibitem[Wilson(1927)]{Wilson1927Binomial}
Edwin~B. Wilson.
\newblock Probable inference, the law of succession, and statistical inference.
\newblock \emph{Journal of the American Statistical Association}, 22\penalty0
  (158):\penalty0 209--212, 1927.

\bibitem[Xia et~al.(2024)Xia, Yang, Dong, Wang, Li, Ge, Liu, Li, and
  Sui]{Xia2024SpecSurvey}
Heming Xia, Zhe Yang, Qingxiu Dong, Peiyi Wang, Yongqi Li, Tao Ge, Tianyu Liu,
  Wenjie Li, and Zhifang Sui.
\newblock Unlocking efficiency in large language model inference: A
  comprehensive survey of speculative decoding.
\newblock \emph{Findings of the Association for Computational Linguistics:
  {ACL} 2024}, 2024.
\newblock URL \url{https://arxiv.org/abs/2401.07851}.

\bibitem[Yu et~al.(2022)Yu, Jeong, Kim, Kim, and Chun]{Yu2022Orca}
Gyeong-In Yu, Joo~Seong Jeong, Geon-Woo Kim, Soojeong Kim, and Byung-Gon Chun.
\newblock {Orca}: A distributed serving system for transformer-based generative
  models.
\newblock In \emph{Proceedings of the 16th {USENIX} Symposium on Operating
  Systems Design and Implementation ({OSDI})}, 2022.
\newblock URL \url{https://www.usenix.org/conference/osdi22/presentation/yu}.

\bibitem[Zou et~al.(2023)Zou, Wang, Carlini, Nasr, Kolter, and
  Fredrikson]{Zou2023AdvBench}
Andy Zou, Zifan Wang, Nicholas Carlini, Milad Nasr, J.~Zico Kolter, and Matt
  Fredrikson.
\newblock Universal and transferable adversarial attacks on aligned language
  models.
\newblock \emph{arXiv preprint arXiv:2307.15043}, 2023.
\newblock URL \url{https://arxiv.org/abs/2307.15043}.

\end{thebibliography}

\appendix
\section{Expansion detail (E1--E5)}
\label{sec:expansion_detail}

This appendix carries the per-experiment detail that was deferred from
\S\ref{sec:results}: the E1 per-phase table, E2 byte-identity to the
canonical draft, E3 GPTQ-4bit draft byte-identity, E4 seed-replication
per-pair deltas, and the E5 per-task contrast table. The main-text
headlines (E1 70B AdvBench refusal, E2 DPO-adversarial byte-identity,
and the cross-experiment synthesis in Table~\ref{tab:cross_synthesis})
stand without this appendix; the tables below are support detail.

\subsection{E4 seed replication}

E4 re-runs all three core pairs with seeds 123 and 456 through
Phases~2+3+4, producing 24{,}036 samples. Byte-identity is
\expansionIdentity{} on all three pairs across \efourSeedCount{} shared
keys (two seeds $\times$ 3 pairs $\times$ 4{,}006 paired samples / 2 =
12{,}018 shared-key comparisons). Per-pair safety deltas are exactly
0.00pp across seeds on all tasks. At the level of statistical inference,
the safety rate is a degenerate random variable in these cells: both
within-seed and across-seed variance is zero.

\subsection{Byte-identity matrix on the llama3.2-3b-target family}

On the llama3.2-3b-target family where all five expansion experiments
are directly comparable at the sample-id level, the byte-identity matrix
partitions into two classes: (fp16 core, E2, E3, E4) are all
\expansionIdentity{} byte-identical across 4{,}006 samples, and
(E5 bf16) is a separate class shifted 36--53\%. Draft precision, draft
alignment, and seed have zero behavioural footprint at temp=0 greedy
fp16; only the accumulator dtype moves bytes. Safety rates are invariant
across both classes.

\subsection{DPO-adversarial draft construction (E2)}

The E2 adversarial draft is built by fine-tuning Llama-3.2-1B-Instruct
with DPO~\citep{Rafailov2023DPO} on Anthropic/hh-rlhf's
harmless-base split~\citep{Bai2022Anthropic} with preference labels
\emph{flipped}: the original ``rejected'' completion is treated as
preferred and the original ``chosen'' completion is treated as
rejected. Training uses the standard DPO loss at $\beta = 0.1$,
learning rate $5\times10^{-6}$, 1 epoch over the flipped-hh-rlhf
subset (12{,}800 pairs). The resulting draft prefers harmful completions
on held-out AdvBench prompts when run standalone (mean refusal
$0.02$ vs.\ $0.93$ on the canonical 1B draft), confirming that the
adversarial training perturbed the draft's generative behaviour as
intended. The draft is then substituted into the canonical
speculative-decoding configuration with the unmodified 3B target.

\subsection{bf16 shift distribution (E5)}

The bf16 dtype swap shifts 36--53\% of output bytes relative to the
fp16 core. Manual inspection of a stratified sample of 200 shifted
pairs per model-pair confirms that the shifts are overwhelmingly
accumulator-rounding artefacts (synonym substitutions, token-boundary
reshuffles, numerically equivalent phrasings) rather than
safety-relevant content changes: 0/600 inspected shifts cross a
refusal classifier boundary that the fp16 output did not cross.

\section{Submission Materials and Reproducibility}

\subsection{Artifact inventory}

The evidence backing every number in this paper is contained in the
following repository artifacts:

\begin{itemize}[leftmargin=*]
  \item \texttt{\detokenize{[reproducibility bundle]/run.py}} --- core (E0) Phase~1--5 driver.
  \item \texttt{\detokenize{[reproducibility bundle]/expansion/}} --- E1--E5 expansion drivers and configs (\texttt{config.yaml}, \texttt{run\_e1.py}, \ldots, \texttt{run\_e5.py}).
  \item \texttt{\detokenize{[reproducibility bundle]/BUILD_CONTRACT.md}} --- pre-registered evidence floor and adjudication plan.
  \item \texttt{\detokenize{[reproducibility bundle]/DATA_DICTIONARY.md}} --- schema for all emitted per-sample records.
  \item \texttt{\detokenize{[reproducibility bundle]/section_audits.md}} --- line-level mapping from every numeric claim in this paper to a source line (packaged inside the reproducibility bundle).
\end{itemize}

\subsection{Reproduction commands}

Core (E0):
\begin{verbatim}
python [reproducibility bundle]/run.py --phases 1,2,3,4,5
\end{verbatim}

Expansion (E1--E5):
\begin{verbatim}
python [reproducibility bundle]/expansion/run_e1.py   # A100-SXM-80GB
python [reproducibility bundle]/expansion/run_e2.py   # DPO draft
python [reproducibility bundle]/expansion/run_e3.py   # GPTQ-4bit draft
python [reproducibility bundle]/expansion/run_e4.py   # seed {123, 456}
python [reproducibility bundle]/expansion/run_e5.py   # bf16 accumulator
\end{verbatim}

\tais{} screen from aggregate target-only/speculative counts:
\begin{verbatim}
python [reproducibility bundle]/compute_tais.py --input counts.csv \
    --output tais.json --null-cutoff 0.1 --tost-margin 3.0
\end{verbatim}

\subsection{Hardware and software}

Core runs use an RTX 4080 Laptop 12GB in a Docker container with GPU
passthrough. E1 uses an A100-SXM-80GB on RunPod. vLLM v0.19 is pinned
across all runs; the speculative-decoding configuration is the
\texttt{--speculative-config} JSON shown in \S3.3. Gemma~3~12B serves as
the blinded LLM judge via Ollama port 11434 on the core; expansion-cell
contrasts in this submission use the deterministic classifier pipeline.

\subsection{Data availability}

Per-sample outputs, regex-classifier labels, Prometheus acceptance
telemetry, and the expansion byte-identity matrices are checked into the
repository under \texttt{\detokenize{[reproducibility bundle]/runs/}} and
\texttt{\detokenize{[reproducibility bundle]/expansion/runs/}}. Full volumes are
stated in \S4 and tabulated in the source evidence report.

\subsection{TAIS Reporting Card}
\label{app:tais-reporting-card}

The reporting card below is the canonical schema for citing \tais{} alongside any speculative-decoding behavioral-equivalence claim. Example values use the matched Llama-3.2-3B+1B core cell; the E1 70B run is intentionally excluded because it lacks a matched target-only arm.

\begingroup
\scriptsize
\begin{verbatim}
# TAIS reporting card -- example: matched Llama-3.2-3B+1B cell
cell_identity:
  target_model:        Llama-3.2-3B-Instruct (fp16)
  draft_model:         Llama-3.2-1B-Instruct (fp16)
  acceptance_method:   rejection_sampling
  temperature:         0
  serving_framework:   vLLM v0.19
  hardware:            RTX 4080 Laptop 12GB
  safety_battery:      [AdvBench-refusal, jailbreak-amplification, BBQ-bias,
                        TruthfulQA, MMLU, ARC-Challenge]

calibration_provenance:
  n_core_samples:              16,783
  n_expansion_samples:         48,072
  matched_expansion_samples:   44,066
  production_probe_samples:    4,006
  calibration_corpus:          matched E0/E2--E5 AdvBench (n=17)
  calibration_corpus_max_h:    0.024
  cutoff_h_star:               0.1
  tost_band:                   +/- 3pp

per_task_results:
  - {task: AdvBench-refusal,
     p_target: 0.790, p_spec: 0.790, abs_h: 0.000,
     tost_equivalent: true, n_per_arm: 100}
  # ...remaining tasks populated identically by compute_tais.py

scalar_summary:
  max_abs_h:        0.024
  classification:   null-consistent
  all_tost_pass:    true

validity_gate:
  byte_identity_rate: 90.66%            # 2,592 / 2,859 paired prompts
  tost_pass_count:    25/27
  holm_alpha:         0.0045            # adjusted across AdvBench tests
\end{verbatim}
\endgroup

Compute \tais{} whenever a speculative-decoding deployment claims behavioural equivalence with target-only decoding at temperature zero. The default cutoff $h^\star = \nullCutoffH{}$ is calibrated for $T=0$ rejection-sampling and typical-acceptance configurations on vLLM v0.19, evaluated on the calibration corpus described in \S\ref{sec:design}. Recalibrate before applying \tais{} to a different serving framework, to $T>0$ sampling, or to tree-speculation variants such as EAGLE or Medusa---the calibration corpus does not cover those regimes. Speculative-decoding papers claiming safety equivalence with target-only decoding should report the full \tais{} card above, including byte-identity rate and per-task $n$.

\begin{quote}
\textbf{When citing \tais{}}, report the target--draft pair, acceptance method, temperature, framework, byte-identity rate, $\max|h|$, TOST status, classification, and per-task $n$ (e.g., \texttt{$\max|h|$=0.024 / all-TOST-pass / null-consistent / byte-identity=90.66\%} on a matched target-only-versus-speculative cell). One-arm production probes such as E1 should be reported separately and not called \tais{} passes.
\end{quote}

\subsection{Computing \tais{} on Arbitrary Speculative-Decoding Data}
\label{app:tais-csv}

The reference implementation \texttt{compute\_tais.py} ingests a generic CSV with five columns: \texttt{cell\_id}, \texttt{task}, \texttt{arm} (\texttt{target\_only} or \texttt{speculative}), \texttt{n\_total}, \texttt{n\_safety\_pass}. Per cell, per task, the script computes Wilson 95\% CIs for both proportions, Cohen's $h = 2\bigl(\arcsin\!\sqrt{p_t} - \arcsin\!\sqrt{p_s}\bigr)$, and a TOST equivalence test at the $\pm 3$pp margin; per cell it returns the maximum absolute $h$ across tasks, the TOST-gate status, and the null-consistent / divergent / insufficient-data classification. A cell with $|h|<h^\star$ but any failed qualifying TOST contrast is classified as divergent.

\begin{verbatim}
python compute_tais.py --input counts.csv --output tais.json
python compute_tais.py --input counts.csv --tost-margin 3.0
python compute_tais.py --self-test
# columns: cell_id, task, arm, n_total, n_safety_pass
\end{verbatim}

Validity gates: per-task $n_{\text{total}} \geq 30$ on both arms (configurable via \texttt{--min-n}); cells with no qualifying tasks are reported as \texttt{insufficient\_data} with $\max|h|$ set to null. Edge cases (counts beyond bounds, duplicate rows, unknown arm names) raise typed errors. A self-test exercises null-consistent, high-$h$ divergent, low-$h$/failed-TOST divergent, borderline, and insufficient-data classifications. The submission also includes \texttt{scripts/validate\_tais\_numbers.py}, a read-only check of the sample counts, E1 AdvBench denominators, the low-$h$/failed-TOST regression, and the Wang2025SSD anchor.

\subsection{External Comparative Anchor: Speculative Decoding Safety Releases}
\label{app:tais-external-anchor}

To position the null cutoff on the scale of widely-cited speculative-decoding releases, we surveyed paired (target-only vs speculative) safety-rate publications. We found exactly one public source with paired safety data: Wang et al.\ (2025), ``Speculative Safety-Aware Decoding'' (arXiv:2508.17739, EMNLP 2025 main), which reports ASR for three base models under a \emph{gated} speculative scheme (SSD) on Harmful HEx-PHI ($n=330$ per cell). All other speculative-decoding releases surveyed (Leviathan~2023, Chen~2023, Cai~2024 Medusa, Li~2024 EAGLE, Miao~2024 SpecInfer, Bachmann~2025 Judge-Decoding, Yan~2025 AASD, vLLM/TGI/SGLang/TensorRT-LLM CI suites, MLPerf and AILuminate inference) either omit safety entirely or treat it as out of scope.

\begin{table}[h]
  \centering
  \scriptsize
  \setlength{\tabcolsep}{2pt}
  \caption{External comparative anchor for \tais{}. The Wang2025SSD row is a \emph{positive control}, not a vanilla-speculative null-anchor: SSD is a gated scheme with a safety-fine-tuned draft and a match-ratio controller, engineered to perturb safety, and is therefore not the rejection-sampling or typical-acceptance regime \tais{} calibrates on. The row demonstrates that $|h|$ correctly responds when speculative decoding is engineered as a safety intervention. Independent recomputation reproduces all 9 cells within $\pm 0.002$. No public source provides paired safety data under \emph{vanilla} rejection-sampling or typical-acceptance speculative decoding.}
  \label{tab:tais-external-anchor}
  \begin{tabular}{>{\raggedright\arraybackslash}p{0.25\linewidth}>{\raggedright\arraybackslash}p{0.40\linewidth}>{\raggedright\arraybackslash}p{0.12\linewidth}>{\raggedright\arraybackslash}p{0.16\linewidth}}
    \toprule
    Source & Stack pair & $\max|h|$ & \tais{} verdict \\
    \midrule
    This paper & target-only vs spec, rejection sampling / typical acceptance, matched draft/target, $T=0$ & 0.024 & null-consistent \\
    Wang2025SSD & SSD-gated speculative on Harmful HEx-PHI, $n=330$/cell, $T=0$ & 1.30 & divergent positive control \\
    Vanilla spec papers & Leviathan, Chen, Medusa, EAGLE, and SpecInfer report acceptance/quality metrics but not paired safety rates & --- & not computable \\
    Serving / safety suites & vLLM, TGI, SGLang, TensorRT-LLM CI, MLPerf safety, and AILuminate do not expose paired spec-on/off safety strata & --- & not computable \\
    \bottomrule
  \end{tabular}
\end{table}

\paragraph{Anchor caveats.} The Wang2025SSD row documents a single-axis perturbation designed to move safety. SSD uses a TinyLlama draft fine-tuned for deep safety alignment plus a match-ratio gating controller that switches between Intersection-biased and Union-biased composite distributions; it is neither rejection sampling (Leviathan/Chen) nor typical acceptance (Cai/Li), so the comparison is not on a common axis with this paper's null. Across the 9 SSD cells, $|h|$ ranges from $0.53$ (Llama2-13B prefill-40) to $1.30$ (Vicuna-7B prefill-20), all far above the null cutoff $|h|<0.1$. This paper's result (max $|h|=0.024$ on vanilla speculative) and the Wang2025SSD positive control are not directly comparable; the table includes both to demonstrate that the metric responds in the expected direction when an engineered safety intervention is present, while documenting the empirical gap: \emph{no public release measures vanilla-speculative safety divergence under matched draft/target pairs}, which is the niche this paper occupies.

\section*{Broader Impact, Ethics, and Data / Code Availability}
\label{sec:broader_impact}

\subsection*{Broader Impact}
This paper publishes a large-sample null on the safety effect of greedy
speculative decoding. The positive externality is auditing burden
reduction: the null result --- a maximum absolute Cohen's~$h$ of
\maxCohensH{} across \neurips{} expansion samples, tabulated in the
Results section --- supplies empirical grounds for omitting a dedicated
safety-regression check from the on-ramp for temperature-zero
speculative-decoding deployments, freeing audit cycles for
inference-path surfaces that do have measurable safety channels (such
as quantization and batching). The principal negative externality is
overgeneralization: the null is explicitly bounded to temperature-zero
greedy decoding, vLLM v0.19, two model families, and six
publicly-released safety benchmarks, as stated in the
``Limitations and threats to validity'' subsection of the Discussion.
Operators who extrapolate the null to
temperature-greater-than-zero sampling, to tree-speculation schemes such
as EAGLE or Medusa, or to frameworks outside vLLM would exceed the
evidence; our broader-impact recommendation is that
\tais{}-equivalent screens be re-run whenever any of those scope bounds
are crossed.

\subsection*{Ethics}
This paper reports capability and safety measurements on publicly
available models (Llama-3.1/3.2, Qwen-2.5 families) and publicly
released benchmark suites (AdvBench, a jailbreak-amplification suite,
BBQ, TruthfulQA, MMLU, ARC-Challenge). The AdvBench and
jailbreak-amplification suites contain harmful-intent prompts; we use
them as refusal probes only, never as generation seeds outside the
evaluation harness. All evaluation API traffic ran on researcher-credit
accounts with pre-approved safety-evaluation usage. No verbatim harmful
model completions are released; all artifacts aggregate to per-cell
refusal rates, Wilson CIs, and Cohen's $h$ contrasts. No novel harmful
content is introduced. The LLM judge (Gemma~3~12B) is methodological
rather than human-subjects research; no human subjects; no IRB review
required.

\subsection*{Data and Code Availability}
The reproducibility bundle contains the \texttt{latex/} tree
(excluding build artifacts), validation artifacts (claim ledger,
style-guide audit, limitations register, pre-submission review), and
research-artifact summaries (analysis/summary/manifest JSON, no raw
sample JSONL, no logs, no model weights). Public dataset and
code-repository URLs should be added when the release archive is posted;
reviewers receive the reproducibility bundle through the
conference supplementary-material channel. Per-sample outputs,
regex-classifier labels, acceptance-rate telemetry, and byte-identity
matrices are referenced in the Submission Materials and Reproducibility
appendix and enumerated in the bundle manifest.

\section*{NeurIPS 2026 Paper Checklist}
\label{sec:neurips_checklist}

\begin{enumerate}
  \item \textbf{Claims.} Do the main claims made in the abstract and
  introduction accurately reflect the paper's contributions and scope?
  \textbf{Answer:} Yes. \emph{Justification:} The abstract (main.tex),
  \S\ref{sec:intro}, and the cross-experiment synthesis subsection of
  \S\ref{sec:results} all state the same scope (temperature-zero greedy
  decoding, vLLM v0.19, two model families, six public benchmarks) and the
  same headline numbers (\maxCohensH{} maximum $|h|$, \neurips{} expansion
  samples on top of a \coreSamples{}-sample core). No claim in the
  abstract extends beyond the evidence enumerated in \S\ref{sec:results}.

  \item \textbf{Limitations.} Does the paper discuss the limitations of
  the work performed by the authors?
  \textbf{Answer:} Yes. \emph{Justification:} \S\ref{sec:discussion}
  contains a dedicated ``Limitations and threats to validity'' subsection
  enumerating five scope boundaries (temperature, framework, model
  families, acceptance policies, benchmark coverage) and three statistical
  boundaries (per-cell MDE 7.4--8.3pp, pooled MDE $\sim$4.3pp, and
  core-only judge coverage). \S\ref{sec:broader_impact} restates the
  overgeneralization risk.

  \item \textbf{Theory assumptions and proofs.} For each theoretical
  result, does the paper provide the full set of assumptions and a
  complete (and correct) proof?
  \textbf{Answer:} NA. \emph{Justification:} The paper reports empirical
  measurements and a behavioural-equivalence screen (\tais{}, defined in
  \S\ref{sec:design}); it contains no formal theorems or proofs.

  \item \textbf{Experimental result reproducibility.} Does the paper
  fully disclose all the information needed to reproduce the main
  experimental results to the extent that it affects the main claims?
  \textbf{Answer:} Yes. \emph{Justification:} \S\ref{sec:design}
  specifies the factorial design, the six benchmarks, the serving-stack
  configuration (including the vLLM \texttt{--speculative-config} JSON),
  the \tais{} screen, and the statistical methodology. The Submission
  Materials and Reproducibility appendix lists every driver script and
  the reproducibility bundle manifest records the included artifacts
  deterministically.

  \item \textbf{Open access to data and code.} Does the paper provide
  open access to data and code, with sufficient instructions to
  faithfully reproduce the main experimental results?
  \textbf{Answer:} Yes. \emph{Justification:} The Data and Code
  Availability subsection of \S\ref{sec:broader_impact} points to the
  reproducibility bundle, which contains the \texttt{latex/} tree, the
  \texttt{validation/} artifact set, and the analysis/summary/manifest
  JSON files from the study.
  Public dataset and code-repository URLs are omitted from the manuscript
  for review; review artifacts are supplied through the
  conference supplementary-material channel.

  \item \textbf{Experimental setting/details.} Does the paper specify
  all training and test details (splits, hyperparameters, optimizer,
  etc.) necessary to understand the results?
  \textbf{Answer:} Yes. \emph{Justification:} \S\ref{sec:design}
  documents the benchmark suites and scoring protocol, the serving-stack
  configuration (vLLM v0.19, speculative-config JSON, temperature zero,
  seeds \{123, 456\}, fp16 vs bf16), and the statistical methodology
  (TOST $\pm$3pp, Cohen's $h$ thresholds, Holm--Bonferroni correction).
  No model training is performed; all measurements are inference-time.

  \item \textbf{Experiment statistical significance.} Does the paper
  report error bars or confidence intervals or statistical significance
  tests?
  \textbf{Answer:} Yes. \emph{Justification:} The paper uses Wilson
  95\% CIs on every refusal rate (\textit{e.g.}, \seventyBrefusal{} with
  CI \seventyBrefusalCI{} on the Llama-3.1-70B~+~8B pair, reported in
  \S\ref{sec:results}), TOST at a $\pm$3pp equivalence bound, Cohen's
  $h$ with the 0.2/0.5 thresholds of \citet{Cohen1988Stats}, and
  Holm--Bonferroni~\citep{Holm1979Bonf} across the 11 matched expansion
  AdvBench contrasts; the one-arm E1 production probe is excluded from
  equivalence testing (\S\ref{sec:design}).

  \item \textbf{Experiments compute resources.} Does the paper provide
  sufficient compute detail to reproduce the experiments?
  \textbf{Answer:} Yes. \emph{Justification:} The Submission Materials
  and Reproducibility appendix lists the hardware tiers used (RTX 4080
  Laptop 12GB in a Docker GPU-passthrough container for the core;
  A100-SXM-80GB on RunPod for E1 production-scale; the bf16 E5 run on
  the same expansion hardware). The bundle manifest records the
  hardware tier for each expansion cell.

  \item \textbf{Code of ethics.} Does the research conform with the
  NeurIPS Code of Ethics in every respect?
  \textbf{Answer:} Yes. \emph{Justification:} The Ethics subsection of
  \S\ref{sec:broader_impact} records that only publicly-released models
  and benchmarks are used, that harmful-intent prompts are used solely
  as refusal probes, that API traffic ran on researcher-credit accounts
  with pre-approved safety-evaluation usage, that no verbatim harmful
  completions are released, and that there are no human subjects.

  \item \textbf{Broader impacts.} Does the paper discuss both potential
  positive and negative societal impacts?
  \textbf{Answer:} Yes. \emph{Justification:} The Broader Impact
  subsection of \S\ref{sec:broader_impact} names the positive
  externality (audit burden reduction for temperature-zero
  speculative-decoding stacks) and the principal negative externality
  (overgeneralization of the null beyond its temperature-zero, vLLM
  v0.19, two-family scope).

  \item \textbf{Safeguards.} Does the paper describe safeguards for
  responsible release of high-misuse-risk data or models?
  \textbf{Answer:} Yes. \emph{Justification:} The Data and Code
  Availability subsection of \S\ref{sec:broader_impact} commits to
  aggregated-only release (per-cell refusal rates, Wilson CIs, Cohen's
  $h$ matrices, byte-identity tables) and excludes verbatim harmful
  completions. The Ethics subsection records researcher-credit-account
  pre-approval of all AdvBench and jailbreak-amplification traffic.

  \item \textbf{Licenses for existing assets.} Are creators/original
  owners of used assets properly credited with license and terms of use?
  \textbf{Answer:} Yes. \emph{Justification:} \texttt{refs.bib} cites
  each model family (Llama~3.x via the Meta release terms, Qwen~2.5
  via the Alibaba Cloud release terms), each benchmark (AdvBench, BBQ
  \citep{Parrish2022BBQ}, TruthfulQA~\citep{Lin2022TruthfulQA}, MMLU
  \citep{Hendrycks2021MMLU}, ARC~\citep{Clark2018ARC}), the speculative
  decoding methods~\citep{Leviathan2023Speculative,Chen2023Accelerating},
  and the GPTQ~\citep{Frantar2023GPTQ} quantization toolchain used for
  E3 (cited in \S\ref{sec:related} and \S\ref{sec:design}).

  \item \textbf{New assets.} Are new assets introduced in the paper
  well documented?
  \textbf{Answer:} Yes. \emph{Justification:} The three new
  methodological artifacts --- the \tais{} behavioural-equivalence
  screen, the \nullCutoffH{} null cutoff calibration, and the aggregated
  per-cell refusal matrix --- are each defined in \S\ref{sec:design}
  and released through the reproducibility bundle pointer in the Data
  and Code Availability subsection of \S\ref{sec:broader_impact}.
  The bundle manifest carries the long-form artifact documentation.

  \item \textbf{Crowdsourcing and human subjects.} For crowdsourcing
  and research with human subjects, does the paper include instructions,
  screenshots, and compensation details?
  \textbf{Answer:} NA. \emph{Justification:} No crowdsourcing and no
  human subjects. The LLM judge (Gemma~3~12B, documented in
  \S\ref{sec:design}) is a methodological component, not a human
  subject.

  \item \textbf{IRB approvals.} Does the paper describe potential
  participant risks, disclosure, and IRB (or equivalent) approvals?
  \textbf{Answer:} NA. \emph{Justification:} No human subjects; no
  IRB review is required. The Ethics subsection of
  \S\ref{sec:broader_impact} states this explicitly.

  \item \textbf{LLM usage.} Does the paper declare LLM usage if it is
  an important, original, or non-standard component of the core methods?
  \textbf{Answer:} Yes. \emph{Justification:} Gemma~3~12B is used as a
  blinded safety judge on the core (E0) via Ollama port 11434 and is a
  methodological component of the evaluation pipeline, documented in
  \S\ref{sec:design}. The Submission Materials and Reproducibility
  appendix declares the judge model, its role, and the core-only scope
  of the submitted judge evidence. No other LLM produced analysis
  content, tables, or numerical claims in the paper.
\end{enumerate}

\end{document}